\documentclass[conference]{IEEEtran}
\IEEEoverridecommandlockouts
\usepackage{cite}
\usepackage{amsmath,amssymb,amsfonts}
\usepackage[utf8]{inputenc} 
\usepackage{algorithmic}
\usepackage{graphicx}
\usepackage{textcomp}
\usepackage{xcolor}
\usepackage{eucal}
\usepackage{url}
\usepackage{algorithm}
\usepackage{algorithmic}
\def\BibTeX{{\rm B\kern-.05em{\sc i\kern-.025em b}\kern-.08em
    T\kern-.1667em\lower.7ex\hbox{E}\kern-.125emX}}
    
\DeclareUnicodeCharacter{2212}{-}
\begin{document}

\title{Towards Stable Imbalanced Data Classification via  Virtual Big Data Projection}

\author{\IEEEauthorblockN{Hadi Mansourifar}
\IEEEauthorblockA{\textit{Computer Science Department} \\
\textit{University Of Houston}\\
Houston, USA \\
hmansourifar@uh.edu}
\and
\IEEEauthorblockN{Weidong Shi}
\IEEEauthorblockA{\textit{Computer Science Department} \\
\textit{University Of Houston}\\
Houston, USA \\
wshi3@uh.edu}

}

\maketitle

\begin{abstract}
Virtual Big Data (VBD) proved to be effective to alleviate mode collapse and vanishing generator gradient as two major problems of Generative Adversarial Neural Networks (GANs) very recently.  In this paper, we investigate the capability of VBD to address two other major challenges in Machine Learning including deep autoencoder training and imbalanced data classification. First, we prove that, VBD can significantly decrease the validation loss of autoencoders via providing them a huge diversified training data which is the key to reach better generalization to minimize the over-fitting problem. Second, we use the VBD to propose the first projection-based method called cross-concatenation to balance the skewed class distributions without over-sampling. We prove that, cross-concatenation can solve uncertainty problem of data driven methods for imbalanced classification.
\end{abstract}


\section{Introduction}
Virtual Big Data (VBD) [44] initially proposed to provide the Generative Adversarial Neural Networks (GANs) [1,2,3,4] with the sufficient training instances when it comes to synthesize extremely scarce positive instances. Soon after, it became clear that, VBD can alleviate mode collapse and vanishing generator gradient problems of the GANs. To generate a virtual instance, $c$ different original training instances are selected and concatenated to each other, where $c$ is concatenation factor. The dimension of output virtual big data is $c*d$ and the size of virtual big data is ${C(N,c)}$, where $N$ is the number of original training instances and $d$ is original dimension. In this paper, we extend the application of VBD to two important Machine Learning issues including deep Autoencoder training and imbalanced data classification. First, we demonstrate how VBD can improve issues related to Autoencoders like dimensionality reduction, anomaly detection and variational data generation. Afterwards, we propose a new method to generate VBD called Cross-Concatenation which can solve the uncertainty and instability of over-sampling techniques with competitive imbalanced classification results.

\subsection{Virtual Big Data (VBD)}
VBD [44] is created by transferring the original data into higher dimension via concatenating the training instances with respect to a concatenation factor. Given a set of training instances $(x_{1}, x_{2},\cdots, x_{w})$, where $x_{i}$ $\in$ $R_{n}$ , $u$ virtual data points are created as $(x'_{1},y_{1}), (x'_{2},y_{2}),\cdots , (x'_{u},y_{u})$, where $x'_{i}  \in R^{cn} , y_i \in R^n $ by concatenation of $c$ selected training instances. The VBD dimension is $c*d$ and the VBD size is ${C(N,c)}$, where $N$ is the number of original training instances and $d$ is original dimension.

\subsection{VBD for GAN-based Data Augmentation}
In recent years, GANs have been used successfully in data augmentation [1,2,3]. The main idea is to train a generative network in adversarial mode against a discriminator network in order to augment minority class by synthesized data. Generative adversarial nets (GANs) have proven to be good at solving many tasks. Since the invention of GAN [4], it has been well used in different machine learning applications [5,6], especially in computer vision and image processing [7,8]. However, GANs need a huge training data to generate efficient augmented data which is not available in many applications.
The curse of dimensionality of VBD might be considered as a negative point. However, it can make the discriminator less perfect and helps the generator to be more competitive by avoiding diminishing gradients. Beyond that, V-GAN, a GAN trained by VBD suffers less mode collapse since each virtual instance contains $c$ different original instances belonging to $c$ different modes. To guarantee the high diversified generated outputs, a diversity maximization function is applied on $c*n$ extracted instances, where $n$ is number of GAN outputs [44].
\subsection{VBD -Based  Deep Autoencoder Training}
Autoencoders [9,10,11,12,13] can learn the deep features of data by transferring the training instances into deep lower dimensions. The unsupervised nature of autoencoders make them perfect to work with huge high dimensional data from various types.  The lack of diversity in training data is the root of autoencoder problems including poor generalization and over-fitting. In this paper, we use Virtual Big Data (VBD) to overcome mentioned problems to train more efficient autoencoders. Comparing to original training data, VBD has two characteristics: larger size and more diversity. These characteristics are what is needed to train an efficient autoencoder. Our experiments show that, autoencoders trained by VBD can not only reach lower validation loss but they can well adapt to anomaly detection tasks. VBD enables us to create several versions of training and test instances by concatenating them with different training instances. As a result, we can evaluate the reconstruction loss of each test data several times by concatenating it to different training instances. Comparing the reconstruction loss of all versions per each test instance with a group of randomly selected training instances enables us to find the outliers with more accuracy. This procedure helps us to build a robust deep anomaly detection, since each test data is evaluated several times.
\subsection{VBD -Based  Imbalanced Data Classification via Cross-Concatenation}

Imbalanced datasets can significantly impact the efficiency of learning systems in various domains including pattern recognition and computer  vision.  Researchers deal with the class imbalanced problem in many real-world applications, such as diabetes detection [14], breast cancer diagnosis and survival prediction [15,16], Parkinson diagnosis [17], bankruptcy prediction [18], credit card fraud detection [19] and default probability prediction [20]. In these applications, the main task is to detect a minority instance. However, standard classifiers are generally inefficient for imbalanced data classification due to low rate occurrence of the minority instances. A standard classifier trains the models with bias toward the majority class which leads to high overall accuracy and poor recall score since a large fraction of minority instances would be labeled as majority instance. Solutions to address the class imbalanced problem fall into two categories: data driven approaches and algorithmic approaches. Data driven techniques [21] aim to balance the class distributions of a dataset before feeding the output into a classification algorithm by either over-sampling or under-sampling the data. When the dataset is highly imbalanced, under-sampling could lead to significant loss of information. In such cases, over-sampling has proven to be more effective for dealing with class imbalanced problem. SMOTE is the most popular over-sampling method due to its simplicity, computational efficiency, and superior performance [37]. However, SMOTE blindly synthesize new data in minority class without considering the majority instances, especially in vicinity regions with majority class. On the other hand, the common problem of SMOTE variations is non-stable results due to their random nature meaning that a unique set of synthesized data and classification results are not guaranteed. In fact, in case of running SMOTE $n$ different times, $n$ different synthesized instances are obtained with $n$ different classification results. In this paper, we propose the first projection-based method called Cross-Concatenation to address imbalanced data classification problems. Cross-Concatenation is based on projecting minority and majority instances into new space wherein the data is discriminated better.  First, given $M$ minority instances, each one is concatenated with $N$ majority instances to form $M * N$ new double size data. Second, given $N$ majority instances, each one is concatenated with $M$ minority instances to form $N * M$ new double size data. This approach to project the data can make two classes balanced. Moreover, it can significantly increase the number of training data. To project the test data into new space, each one is concatenated with the centroid of majority and minority classes to form two different instances. To assign the label to each test data, two obtained data corresponding to each test data are passed to trained model and the highest probability returned from the model is used as a metric to assign the label. Our experimental results show that, generating VBD using Cross-Concatenation can balance the data with stable competitive results comparing to traditional over-sampling techniques.
\subsection{Contributions}
In this section we summarize our contributions as follows.

\subsubsection{Deep Autoencoder Training}
\begin{itemize}
    \item We study the application of Virtual Big Data (VBD) to train more efficient autoencoders.
    \item We propose a new method for anomaly detection based on VBD.
    \item We prove that, VBD can significantly decrease the validation loss of Variational Autoencoders.
    \item Our experiments show that, the proposed method can improve anomaly detection results.
    
\end{itemize}

\subsubsection{Imbalanced Data Classification}

\begin{itemize}
    \item We propose the first projection-based method to balance the skewed class distributions using VBD.
    \item The proposed method can solve instability problem of data driven methods for imbalanced classification.  
    \item We show the superiority of proposed method versus SMOTE as the most popular over-sampling technique.
\end{itemize}

The rest of the paper is organized as follows. Section 2 reviews the required backgrounds and related works. Section 3, presents the VBD-based deep autoencoder training. Section 4 introduces 
VBD-based imbalanced data classification. Section  5 demonstrates the experiments. Finally section 6 concludes the paper.

\section{Background and Related Works}
In this section, we briefly review the required concepts and previous researches related to different aspects of autoencoders.

\subsection{Autoencoders}
Autoencoders were proposed by Hinton and the PDP group [22] to address the neural network based unsupervised learning. Autoencoders resurfaced once again with the surge of deep learning [23][24]. A deep autoencoder is a feed-forward multi-layer neural network which maps the output to the input itself [25]. The dimensionality reduction which happens at the bottleneck doesn’t allow this map to be identical. In the other word, autoencoders learn a map from the input to low dimensional version of it and then to itself through a pair of encoding and decoding hidden layers.
$\overline{X}=D(E(X))$, 
where $X$ is the input data, $E$ is an encoding function from the input data to the hidden bottleneck, $D$ is a decoding function from the hidden bottleneck to the output, layer, and $\overline{X}$ is the reconstructed version of the input data. The mission of autoencoder is to train $E$ and $D$ to minimize the difference between $X$ and $\overline{X}$ as follows. \\
\centerline{$\overset{min}{D,E} \left \|  X- D(E(X)) \right \|$}
\subsection{Deep Anomaly Detection}
Anomaly detection is a traditional topic in machine learning [26,27,28,29]. Unsupervised anomaly detection aims to discriminate anomalous data with unknown labels. Deep autoencoders have been used successfully for anomaly detection [28,29]. Robust Deep Autoencoder (RDA) or Robust Deep Convolutional Autoencoder (RCAE) decomposes input data $X$ into two parts $X = L_D + S$, where $L_D$ represents the latent representation of the autoencoder. The matrix $S$ captures the noise and outliers which are hard to reconstruct as shown in Equation 1 [30]. The decomposition is carried out by optimizing the objective function shown in Equation 1.
\begin{equation}
     \overset{min}{\theta ,s} + \left \|  L_D- D_\theta(E_\theta(L_D)) \right \|_2 + \lambda \cdot \left \| S ^T\right \| _{2,1}
\end{equation}

\begin{equation}
s.t \quad X - L_D - S=0
\end{equation}
The above optimization problem is solved using a combination of backpropagation and Alternating Direction Method of Multipliers (ADMM) approach [31].

\subsection{Variational Autoencoders}
Varaiational Autoencoders (VAEs) [33,34,35] are kind of generative models proposed by Kingma and Welling [32]. In addition to traditional encode-decode ability, VAEs have the ability to capture the distribution of the latent vector $z$, which can be considered as independent unit Gaussian random variables, where $z \sim \mathcal{N} (0,I)$. To minimize the difference between distribution of $q(z\mid x)$ and KL Divergence which is a Gaussian distribution, the gradient descent algorithm is used. It allows VAE models to be trained by optimizing the reconstruction loss $(\mathcal {L}_{rec})$ and KL divergence loss $(\mathcal {L}_{ki})$ as follows.
\begin{equation}
    \mathcal {L}_{rec}= - \mathbb{E}_{q (z\mid x)} \left [ log ( p(x \mid z) \right) ]
\end{equation}

\begin{equation}
    \mathcal {L}_{kl}=  D_{kl} ( q(z \mid x) \parallel p(z))
\end{equation}

\begin{equation}
    \mathcal {L}_{vae}=\mathcal {L}_{kl} +\mathcal {L}_{rec}
\end{equation}

\subsection{Over-sampling methods}
SMOTE [36] is the most popular over-sampling method due to its simplicity, computational efficiency, and superior performance [37]. However, SMOTE blindly synthesize new data in minority class without considering the majority instances, especially in vicinity regions with majority class. To address this problem, Han et al. proposed Borderline-SMOTE [38], which focuses only on borderline instances in the majority class vicinity regions. However, the precision rate can be highly impacted because classifier fails to detect instances belonging to majority class. Although a superior over-sampling method should ideally improve the minority class detection rate, it must not lead to disability to detect majority instances. To solve this problem, [39] proposed MWMOTE a two-step weighted approach that extends Borderline-SMOTE and ADASYN using the information of the majority instances that lie close to the borderline. Also, [40] proposed DBSMOTE which uses DBSCAN to evaluate the density of each region and then over-samples inside each region to avoid synthesizing an instance inside majority class. A-SUWO [41], is also a clustering-based method designed to identify groups of minority samples that are not overlapped with clusters from the majority class. However, it underestimates the role of noise or mislabeled datapoints which makes it hard to find non-overlapping regions. To Address this problem, CURE-SMOTE [42] proposed denoising and removing outliers before over-sampling.

\subsubsection{SMOTE}
Synthetic Minority Over-sampling Technique (SMOTE) [36] is a method of generating new instances using existing ones from rare or minority class. SMOTE has two main steps: First, the neighborhood of each instance is defined using the $k$ nearest neighbors of each one and Euclidean norm as the distance metric. Next, $N < k$ instances of the neighborhood are randomly chosen and used to construct new samples via interpolation. Given a sample $x_{i}$ from the minority class, and $N$ randomly chosen samples from its neighborhood $x_i^p$,  with $p = 1, . . ., N,$ a new synthetic sample $x_i^{*p}$ is obtained as follows:\\
$x_i^{*p}: = x_{i} + u (x_i^p - x_{i})$ \\
where $u$ is a randomly chosen number between 0 and 1. As a result, SMOTE works by adding any points that slightly move existing instances around its neighbors. To some extent, SMOTE is similar to random over-sampling. However, it does not create the redundant instances to avoid the disadvantage of overfitting. It synthesizes a new instance by random selection and combination of existing instances.
\section{VBD-Based Deep Autoencoder Training}
Given a set of training instances $(x_{1}, x_{2},\cdots, x_{w})$, where $x_{i}$ $\in$ $R_{n}$ , $u$ virtual data points are created as $(x'_{1},y_{1}), (x'_{2},y_{2}),\cdots , (x'_{u},y_{u})$, where $x'_{i}  \in R^{cn} , y_i \in R^n $ by concatenation of $c$ selected original instances. The proposed Algorithm in [44] is well suited for large datasets. To deal with small datasets we use a different algorithm with $c=2$ as follows.

\begin{algorithm}[]
\caption{  Virtual Big Data Synthesis for Small Datasets.
\textbf{Input}: Training instances \textbf{u}; \textbf{Output : }Virtual Big Data \textbf{ v}.}
\begin{algorithmic}[1]
   \STATE {$c_1 \leftarrow 0$}
   
   \STATE {$c_3 \leftarrow 0$}
   
   \FOR{ the size of \textbf{u}}
   \STATE {$c_2 \leftarrow 0$}
    \FOR{ the size of \textbf{u}}
    \STATE{$v_{c_3} \leftarrow  Concatenate (u_{c1} , u_{c2})$}
     
     \STATE {$c_2 \leftarrow c_2+1$}
     \STATE {$c_3 \leftarrow c_3+1$}
     
    \ENDFOR
    \STATE {$c_1 \leftarrow c_1+1$}
     \ENDFOR
     \STATE{Return \textbf{ v} }
\end{algorithmic}
\end{algorithm}

Algorithm 1 can generate $n^2$ instances as Virtual Big Data, where $n$ is the size of original training data . Following algorithm is used to generate specified number of Virtual Big Data [44].
\begin{algorithm}[]
\caption{  Virtual Big Data Synthesis for Large Datasets (X,c,u)
\textbf{Input}: Original instances \textbf{X}; Concatenation factor \textbf{c}; The size of virtual data  \textbf{u}; \textbf{Output}:Virtual Big data \textbf{V}.}
\begin{algorithmic}[1]

   \STATE{ Given \textbf{X}, select \textbf{c} different data points from \textbf{X} and save them in \textbf{U}.}
    \STATE{ Concatenate the instances in \textbf{U} and insert the result to \textbf{V}.}
     \STATE{ If Size(\textbf{V}) $<$ \textbf{u} then Go to Step 1. }
     
    \STATE{Return V.}
      
\end{algorithmic}
\end{algorithm}

The VBD size is significantly larger than original training data size because vector concatenation lets us to increase the size of training data from $n$ to $n^2$ based on Algorithm 1. We first prove that, vector concatenation can significantly increase the efficiency of deep autoencoders by increasing the size of training data. The reason is that, the nearest neighbor of $X$ converges almost surely to $X$
as the training size grows to infinity [44].\\

\textbf{Lemma 1.}  \textit{ (Theorem Convergence of Nearest Neighbor) If $X_1, X_2,..., X_n$ are i.i.d.
in a separable metric space  \scalebox{1.3}{$\chi$} , $X'_n (X)$ $_{=}^{\Delta} X'_n$ is the closest of the $X_1, X_2,..., X_n$ to X in a metric $D(\cdot,\cdot)$, then} \\
$X'_n \rightarrow X$ \quad \textit{a.s.} \\
\textit{Proof.} \quad Let $B_r (x) $ be the (closed) ball of radius r centered at x \\
\begin{equation}
B_r (x) \quad _{=}^{\Delta} \quad \big\{z \in \mathbb{R} ^d
: D(z, x)  \leq r \big \}
\end{equation}
\\
for any $r > 0 $ \\
\begin{equation}
P(B_r (x)) \quad _{=}^{\Delta} \quad P_r \big [ z \in B_r (x) \big ] =  \int _ {B_r(x)} p(z)d(z) > 0  
\end{equation}
Then, for any $\delta > 0$ , we have \\
\begin{equation}
P_{r} \Big [ \underset{i=0,1,2,...}{min} \big \{ D(x_i
, x)\geq \delta \big \}  \Big ] = \Big  [ 1 - P(B_r (x)) \Big]^n  \rightarrow 0 
\end{equation}
\\
There exists a rational
point $a_{\Bar{x}}$ such that $a_{\Bar{x}} \in B_{\bar{r}/3}  (\bar{x})$. Where for some $\bar{r}$, we have $P(B_{\bar{r}}  (\bar{x})) = 0$

Consequently, there exists a small sphere $B_{\bar{r}/2}  (\bar{x})$ such that
\begin{equation}
B_{\bar{r}/2}  (\bar{x}) \subset B_{\bar{r}}  (\bar{x})
\Rightarrow P(B_{\bar{r}/2}  (\bar{x})) = 0 
\end{equation}
\\
Also, $ \bar{x} \in B_{\bar{r}/2}  (\bar{a \bar{x}})$. Since $a \bar{x}$ is rational, there is at most a countable set of such spheres that contain the
entire $ \bar{X}$ therefore, \\
 \begin{equation}
 \bar X \subset  \underset{\bar {x} \in \bar {X}}{\bigcup} 
 B_{\bar{r}/2}  (\bar{a \bar{x}})
 \end{equation}
 \\
and from (7,9) this means $P(\bar{X})$ = 0. Therefore, increasing the size of training set can significantly improve of the classification results.
\subsection{Diversity Measure}
In this section we prove that, the diversity of data is increased by transferring the original instances to VBD. The following sections will introduce the measurements and required discussions [44].
\subsubsection{Distance-based measurements}
Euclidean distance is the simplest way to measure the diversity in a dataset . Generally, datasets with large distances between
different data points show less redundancy. Therefore, enlarging the distances can decrease the similarity and as a result the data can be diversified. Based on Euclidean distance formula \\ \centerline{$\sum\limits_{i \ne j}^{n} || w_i - w_j|| ^ 2 $}

We can prove that, VBD have more average distance from each other. Suppose that, $o=\big\{A,B \big\}$ is our dataset with only two data points.
The corresponding VBD is $v=\big\{v_1,v_2,v_3,v_4\big\}$, where $v_1=(A \frown B) $ , $v_2=(B \frown A) $ , $v_3=(A \frown A) $ and $v_4=(B \frown B) $.
 The Euclidean distance between $A,B$ is represented as 
 $d(A,B) = \sqrt{\sum\limits_{i=1}^{n} (A_{i} - B_{i})^2}$
It's easy to show that, the Euclidean distance between $v_1$ and $v_2$ is
$d(v_1,v_2) = \sqrt{\sum\limits_{i=1}^{n} (A_{i} - B_{i})^2 + (B_{i} - A_{i})^2} = \sqrt{2} \sqrt{\sum\limits_{i=1}^{n} (A_{i} - B_{i})^2}$ \\
 Generally, the maximum distance between concatenated data points is $\sqrt{2} d_o$, where $d_o$ is maximum distance between data points in original data. Larger distances between the vectors in VBD proves higher diversity comparing to original data and as a result, the VBD can provide deep autoencoders more diversified training data to train a model less exposed to over-fitting.
\subsection{Dimensionality Reduction Using VBD}
As we mentioned earlier, we use $c=2$ as concatenation factor. It means that, the VBD dimensions are double size as shown in Table 1. It may raise a serious concern regarding dimensionality reduction as one of the most important applications of autoencoders, since the size of bottleneck is double in autoencoders trained by VBD. However, we can show that, it's not a great deal. Suppose $v$ as an instance of VBD, where $v=A \frown B$. If we pass this instance to an autoencoder 
$v'=Decoder(v)$, where $v'= (A' \frown B')$. As a result, $A'$ and $B'$ can be considered as decoded versions as follows: $A'=Decoder(A)$ and $B'=Decoder(B)$ with the same size as regular decoded instances. That's how dimensionality reduction is done by the autoencoders trained by VBD.

\subsection{Deep Anomaly Detection Using VBD}
Setting the threshold to detect the anomalies based on reconstruction error is a cumbersome task. In this section, we propose a new deep anomaly detection method based on VBD which can simplify this process. Generating VBD lets us to create several instances per each test data. As a result, we can evaluate the reconstruction error several times per each test data. It enables us to devise a more straightforward threshold to find the anomalies. First, we concatenate a test instance with $u$ different training instances. Second, we select $u$ different training data pairs. If different versions of test data fail to reach lower reconstruction error in $w$ cases comparing to randomly selected training pairs, we can consider the corresponding test data as anomaly. If we fix the $u$ , we can manipulate $w$ as a positive integer threshold which is more convenient than traditional thresholds to detect the outliers given the reconstruction errors. Algorithm 3 shows the required steps.

\begin{algorithm}[]
\caption{  Deep Anomaly Detection Using Virtual Big Data
\textbf{Input}: Training instances \textbf{X}; Test Instance  \textbf{T} \textbf{Output : }Anomaly Status \textbf{ $S$}.}
\begin{algorithmic}[1]

   \STATE{Randomly select $u$ different training instances and save them in $P$}
   \STATE{Randomly select $u$ different training instances and save them in $Q$}
   \STATE{$c \leftarrow 0$}
   \FOR{i = 0 ; i++ ; i $<$  \textit{u}}
    
    \STATE{$v \leftarrow  Concatenate (T , P_{i})$}
     \STATE{$z \leftarrow  Concatenate (P_{i}$ , $Q_{i})$}
     \STATE {$p=Reconstruction \_ Error(Autoencoder (v))$}
     \STATE {$q=Reconstruction\_ Error(Autoencoder (z))$}
     \STATE {If \quad ($p > q$) \quad Then}
     \STATE {$\quad \quad c \leftarrow c+1$}
    \ENDFOR

     \STATE{If \quad ($c>w$) \quad  Then }
      \STATE { \quad \quad $s=true$}

     \STATE{Return \textbf{ $s$} }
\end{algorithmic}
\end{algorithm}
\section{VBD-Based Imbalanced Classification}
In this section, we propose a novel method called Cross-Concatenation to produce VBD which works based on data projection to balance skewed class distributions. 
\subsection{Cross-Concatenation}
Given a set of minority instances $x= (u_{1}, u_{2},\cdots, u_{m})$ and majority instances $y= (v_{1}, v_{2},\cdots, v_{n})$, where $u_{i}, v_{j}$ $\in$ $\mathbb{R}{^n}$ , the Cross-Concatenation is defined as follows. $\forall _{u} \forall _v  : u'= (u_i \frown  v_{j}) , v'= (v_j \frown  u_{i})  $, where $u'$ and $v'$ are minority and majority instances projected to new space, respectively and $u'_{i}, v'_{j}$ $\in$ $\mathbb{R}^{2n}$.
Algorithm 4 shows the required steps to project the minority and majority instances into higher dimension.
\begin{algorithm}[]
\caption{  Cross Concatenation (u,v)
\textbf{Input}: Minority instances \textbf{u}, Majority instances \textbf{v}; \textbf{Output : }Projected minority and majority instances \textbf{ $u ^ \prime$} , \textbf{$v ^ \prime$} .}
\begin{algorithmic}[1]
   \STATE {$c_1 \leftarrow 0$}
   \STATE {$c_2 \leftarrow 0$}
   \FOR{ the size of \textit{u}}
    \FOR{ the size of \textit{v}}
    \STATE{$u'_{c_1} \leftarrow  Concatenate (u_i , v_j)$}
     \STATE{$v'_{c_2} \leftarrow  Concatenate (v_j , u_i)$}
     \STATE {$c_1 \leftarrow c_1+1$}
     \STATE {$c_2 \leftarrow c_2+1$}

    \ENDFOR
     \ENDFOR
     \STATE{Return \textbf{ $u ^ \prime$} , \textbf{$v ^ \prime$}}
\end{algorithmic}
\end{algorithm}

\begin{figure}[]
\centering
  \includegraphics[width=70mm]{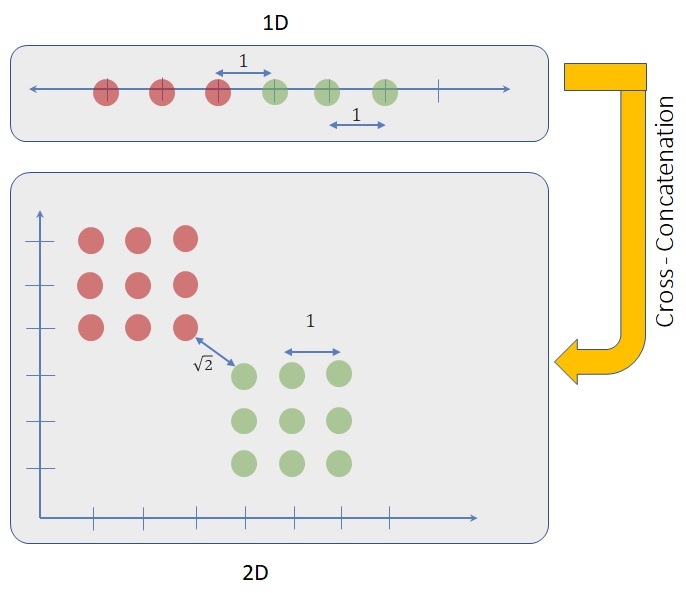}
  \caption{Cross-Concatenation of 1D data into 2D space. Minimum distance of two class instances is increased while minimum distance inside each class is not changed.}
  \label{ }
\end{figure}

\subsubsection{Compactness and Separation}
Since the pairwise distances in high dimensional space are concentrated in limited area, it appears that, high-dimensional spaces are almost empty and it should be easier to separate the classes perfectly with a hyperplane.  However, it's easy to prove that instances are concentrated at the edge of boundary which makes prediction much more difficult. Assume volume of the ball of radius $r$ regarding to dimension $d$ is

\begin{equation}
    V_d (r) = r^d (\frac{\pi ^ {\frac{d}{2}}}{\Gamma (\frac{d}{2} +1}) 
\end{equation}
As a consequence, to cover $[0,1]^d$ with a union of $n$ unit balls we need 
\begin{equation}
   n  \geqslant \frac{1}{V_d} = \frac{\Gamma (\frac{d}{2} +1)}{\pi ^ {\frac{d}{2}}} \quad \underset  {d \rightarrow \infty }{\backsim}  ( \frac{d}{2 \pi e}  ) ^ {\frac{d}{2}} \sqrt{d \pi}
\end{equation}
As a result, if we draw $n$ samples with uniform law in the hypercube, most sample points will be in corners of the hypercube. For example, the probability that a uniform variable on the unit sphere belongs to the shell between the spheres of radius 0.9 and 1 is \\
\begin{equation}
    P(X \in S_{0.9} (p)) = 1- 0.9 ^ d  \quad \underset  {d \rightarrow{\infty}}  { \longrightarrow 1 } 
\end{equation}

Therefore, as dimension increases, the compactness increases and separation decreases. However, the projected data by Cross-Concatenation show completely reverse behaviour. As proved earlier, the maximum distance between data points is $\sqrt{2} d_o$, where $d_o$ is maximum distance between data points in original data. The minimum distance between minority and majority instances also follows this rule as shown in Figure 1. It's a clear sign that projecting the data using the Cross-Concatenation increases the separation between two classes. As a result, the minimum distance in original space is remained unchanged and maximum distance increases which means larger margins. Also, we can prove that, Cross-Concatenation leads to lower VC-dimension due to larger margins. 
Let $S$ be a sample from $\chi$ , such that $\left \| x_i \right \|_2 \leq r$ for all $x_i$ in $S$ then let us define a hypothesis class [33].
\begin{equation}
   \mathcal{H}_{S,\Lambda} \triangleq \begin{Bmatrix}
h_w(x) = sign(w \cdot x) :  _{x_{i} \in S }^{\min} \left | w \cdot x_i \right | = 1 \wedge \left \| w \right \|_2\leqslant \Lambda 
\end{Bmatrix} 
\end{equation}
\textbf{Theorem 1.} \textit{The VC-dimension of $\mathcal{H} _{ S,\Lambda}$ is less than or equal to $r^2 \Lambda ^
2$
.} 
\textbf{Proof}: [33] Let $d$ be the VC-dimension of $\mathcal{H}_{S,\Lambda}$ then there exists
$\begin{Bmatrix}
x_1, ... , x_d
\end{Bmatrix} \subseteq S $
that are shattered by$\mathcal{H}_{S,\Lambda}$ 
Now, we know that for any labeling $(y_1, . . . , y_d)$ there exists
$h_w \in \mathcal{H}_{S,\Lambda} $ such that $y_{i}(w \cdot x_{i}) \geq  1$ for all $i = 1, . . . , d$ . \\
Summing across $i$, we have \\
$d \leq w \cdot \left ( \sum_{i=1}^{d} y_i x_i \right )$

$\leq \left \| w \right \| _2  \left \| \sum_{i=1}^{d} y_i x_i \right \|_2$ \quad (Cauchy-Schwarz inequality) \\
$\leq \left \| \Lambda \right \| _2  \left \| \sum_{i=1}^{d} y_i x_i \right \|_2$ \quad \quad (definition of $\mathcal{H}_{S,\Lambda}$) \\
Now, let $(y_1, . . . , y_d)$ be independent and random with equal probability of 1 or $-1$, so that, $\mathbb{E\begin{bmatrix}
y_i
\end{bmatrix}}=0.$ Now,
if we take the expectation of both sides of the bound, the inequality will still hold. \\
$d \leq \Lambda \mathbb{E\begin{pmatrix}
\left \| \sum_{i=1}^{d} y_i x_i\right \|_2^2
\end{pmatrix} } ^{1/2}$ \\
$ \leq \Lambda \begin{pmatrix} \mathbb{E}
\left \| \sum_{i=1}^{d} y_i x_i\right \|_2^2
\end{pmatrix}  ^{1/2}$ \quad  (Jensen’s inequality) \\

$\leq \Lambda \begin{pmatrix} \mathbb{E}
\left \| \sum_{i,j=1}^{d} y_i y_j x_i x_j\right \|_2^2
\end{pmatrix}  ^{1/2}$ 

$ = \Lambda \begin{pmatrix} 
\left \| \sum _{i,j=1}^{d} \mathbb{E} \begin{bmatrix}
 y_i y_j
\end{bmatrix}   (x_i \cdot x_j)\right \|
\end{pmatrix}  ^{1/2}$ 

$=\Lambda \begin{pmatrix}
\sum_{i}^{d} x_i x_i
\end{pmatrix}^{1/2} \quad (\mathbb{E}\begin{bmatrix}
y_iy_i\end{bmatrix}=1  ,( \mathbb{E}\begin{bmatrix}
y_iy_{i\neq j }\end{bmatrix}=0 ) $ \\
\begin{equation}
   \leq rd ^{1/2} (\left \| x_i  \right \|_2 \leq r) ]] 
\end{equation}
 \\
 So, $d \leq  r^2 \Lambda ^2$.
 Thus , as margin becomes larger, the VC-dimension is decreased
significantly.
We can also prove that there is a relationship between VC-dimension and generalization error. \\
\textbf{Lemma 2.}  [43]. \textit{Suppose $VCD(\mathcal{H}) = d < \infty $. Define }
$\Pi _ {\mathcal{H} } (m) = max \begin{Bmatrix}
\begin{vmatrix}
\Pi _ {\mathcal{H}  }\mathcal{S}
\end{vmatrix}
: \mathcal{S} \subseteq \Omega , |\mathcal{S}=m|
\end{Bmatrix}$ \\

Where i.e.,$\Pi _ {\mathcal{H}  } (m)$ is the maximum size of a projection of $\mathcal{H}$ on an m-subset of $\Omega$) Then \\ 
$\Pi _ {\mathcal{H} } (m) \leq \Phi _d(m) :=\sum_{i=0}^{d} \binom{m}{d} \leq \begin{pmatrix}
\frac{em}{d} 
\end{pmatrix} ^d = O(m^d) $
\\ 

(Note that, if $VCD(\mathcal{H}) = \infty $, then $\Pi _ {\mathcal{H} } (m) = 2^m, \forall m)$ \\

\textit{Proof.} [43] We induct on $m + d$. For $h \in \mathcal{H}$, define $h_\mathcal{S} = h \cap S$. The $m = 0$
and $d = 0$ cases are trivial. Now, consider $m > 0, d > 0$. Fix an arbitrary element $s \in S$. Define \\
$\mathcal{H'}=\begin{Bmatrix}
h_s \in \Pi _ {\mathcal{H} } (\mathcal{S}) | s \notin h_\mathcal{S} , h_\mathcal{S}\cup \begin{Bmatrix}
s
\end{Bmatrix} \in \Pi _ {\mathcal{H} } (\mathcal{S}) 
\end{Bmatrix}$

$|\Pi _ {\mathcal{H} } (\mathcal{S})| = |\Pi _ {\mathcal{H} } (\mathcal{S}- \begin{Bmatrix}
s
\end{Bmatrix})| + |\mathcal{H'}| = |\Pi _ {\mathcal{H} } (\mathcal{S} - \begin{Bmatrix}
s
\end{Bmatrix})| + |\Pi _ {\mathcal{H'} } (\mathcal{S})|$ \\

Since $VCD(\mathcal{H'}
) \leq d - 1$, by induction we obtain \\
$|\Pi _ {\mathcal{H} } (\mathcal{S})| \leq \Phi _d (m-1) + \Phi _{d-1} (m) = \Phi _d (m)$ \\

Thus, we get the following high confidence bound on the generalization error of a function learned from a
function class $\mathcal{H}$ of finite VC-dimension: \\

\textbf{Corollary 1.} [33] Let $\mathcal{H} \subseteq  \big\{−1, 1 \big\} ^ \mathcal{X}$ with $VCD(\mathcal{H}) = d < \infty$. Let $D$ be any distribution on $\mathcal{X} \times \big\{−1, 1 \big\}$,
and let $0 <  \delta < 1$. For any algorithm that given a training sample $\mathcal{S}$ returns a function $h_\mathcal{S} \in \mathcal{H}$, we have
with probability at least $1 - \delta$ over the draw of $\mathcal{S} \sim  D^m $: \\

\begin{equation}
    er_D\begin{bmatrix}
h_\mathcal{S}
\end{bmatrix} \leq er_\mathcal{S} \begin{bmatrix}
h_\mathcal{S}
\end{bmatrix}+ \sqrt{\frac{8(d(\ln(2m)+1)+\ln (\frac{4}{\delta}))}{m}}
\end{equation}

By ignoring constant factors and focusing more on the VC-dimension $(d)$, the number of training
examples $m$, and the confidence parameter $\delta$ we can  write the above bound as \\
\begin{equation}
    er_D\begin{bmatrix}
h_\mathcal{S}
\end{bmatrix} \leq er_\mathcal{S} \begin{bmatrix}
h_\mathcal{S}
\end{bmatrix}+ c\sqrt{\frac{d \ln m + \ln(\frac{1}{\delta})}{m}}
\end{equation}
So, we can argue that, less VC-dimension means less generalization error. That's how Cross-Concatenation makes larger margins, lower VC-dimension and as a result less generalization error.

\subsection{Classification}
Given a set of test instances $t= (t_{1}, t_{2},\cdots, t_{m})$, where $t_{i}$ $\in$ $R{^n}$ , the Cross-Concatenation for test data is defined as follows. $\forall _{t}  : w= (t_i \frown  c_u) , z= (t_i \frown  c_v)  $, where $c_u$ and $c_v$ are centroid of minority and majority classes, respectively and $w_{i}, z_{j}$ $\in$ $R^{2n}$. \\ \\
$\left\{\begin{matrix}
if  \quad P_r[w_i] > P_r[z_i] \quad then  \quad y_i=0
 & \\ 
else  \quad    y_i=1
 & 
\end{matrix}\right.$ \\
Where $P_r[w_i]$ and $P_r[z_i]$ are prediction probabilities returned by the classifier given $w_i$ and $z_i$, respectively.

\begin{table*}[]
\small
\centering
\caption{The structure of tested autoencoders.}
\begin{tabular}{c|c|c|}
\cline{2-3}
                                         & \multicolumn{2}{c|}{\textbf{Hidden Layers}}                                   \\ \cline{2-3} 
                                         & \textit{\textbf{Original Training Data}} & \textit{\textbf{Virtual Big Data}} \\ \hline
\multicolumn{1}{|c|}{\textbf{WBC}}       & 9,6,4,3,4,6,9                            & 18,12,8,6,8,12,18                  \\ \hline
\multicolumn{1}{|c|}{\textbf{Pima}}      & 8,6,4,3,4,6,8                            & 16,12,8,6,8,12,16                  \\ \hline
\multicolumn{1}{|c|}{\textbf{Haberman}}  & 3,2,1,2,3                                & 6,4,2,4,6                          \\ \hline
\multicolumn{1}{|c|}{\textbf{Blood}}     & 4,3,2,3,4                                & 8,6,4,6,8                          \\ \hline
\multicolumn{1}{|c|}{\textbf{Parkinson}} & 22,18,12,6,12,18,22                      & 44,36,24,12,24,36,44               \\ \hline
\multicolumn{1}{|c|}{\textbf{MNIST}}     & 784,128,64,32,64,128,784                 & 1568,256,128,64,128,256,1568       \\ \hline
\multicolumn{1}{|c|}{\textbf{Fashion-MNIST}}     & 784,128,64,32,64,128,784                 & 1568,256,128,64,128,256,1568       \\ \hline
\end{tabular}
\end{table*}

\section{Experiments}
In this section, we test the capability of VBD to enhance the efficiency of deep autoencoder (DAE) training and imbalanced data classification.

\subsubsection{VBD for DAE Experiments}
Our experiments are divided into different parts in order to test the capability of VBD to improve the efficiency of autoencoders. First, we compare the validation loss of autoencoders using the original training data and VBD. Second, we test if VBD can help to enhance the deep anomaly detection results. Finally, we compare the impact of VBD on Variational Autoencoders.

\subsubsection{Datasets}
We used five benchmark datasets available in UCI data repository [13,14,15,16] including Pima Diabity, Wisconsin Breast Cancer, Haberman Survival, Parkinson and Blood Transfusion datasets. In addition, we used two image datasets including MNIST and Fashion-MNIST. These datasets are significantly different in terms of features and attributes.
\subsection{Training-Test Transfer to VBD}
To test the VBD on different autoencoders we need to transfer the original training-test data to their corresponding VBD with the same algorithm. In all test cases except MNIST and Fashion-MNIST, we used Algorithm 1 which generates $n^2$ VBD, where $n$ is the original data size. In case of MNIST datasets, we used Algorithm 2 which were proposed in [44]. 
\begin{figure*}[]
\centering
  \includegraphics[width=150mm]{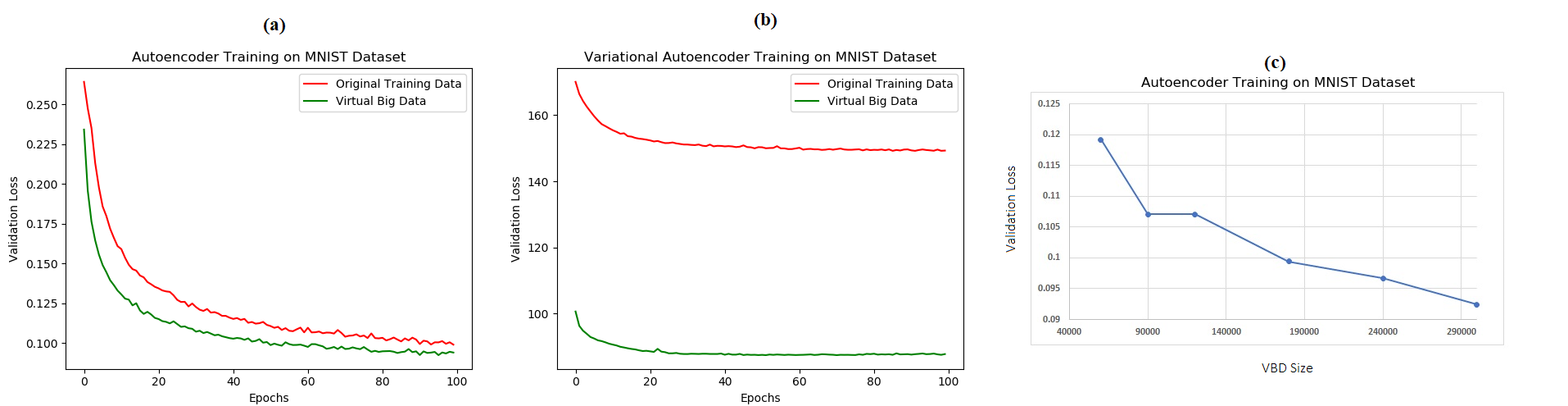}
  \caption{ (a),(b) : Autoencoder training; Original training data versus Virtual Big Data. (c) : The impact of VBD size on validation loss.  }
  \label{ }
\end{figure*}

\begin{table}[]
\centering
\small
\caption{Original training data versus VBD : The comparison of Autoencoder validation losses.}
\begin{tabular}{c|c|c|}
\cline{2-3}
                    & \multicolumn{2}{c|}{\textbf{Validation Loss}}                                 \\ \cline{2-3} 
                                         & \textit{\textbf{Original Training Data}} & \textit{\textbf{Virtual Big Data}} \\ \hline
\multicolumn{1}{|c|}{\textbf{WBC}}       & 0.8467                                   & 0.5458                             \\ \hline
\multicolumn{1}{|c|}{\textbf{Pima}}      & 0.8523                                   & 0.6387                             \\ \hline
\multicolumn{1}{|c|}{\textbf{Haberman}}  & 0.4637                                   & 0.4448                             \\ \hline
\multicolumn{1}{|c|}{\textbf{Blood}}     & 1.0577                                   & 0.8099                             \\ \hline
\multicolumn{1}{|c|}{\textbf{Parkinson}} & 0.7244                                   & 0.5554                             \\ \hline
\multicolumn{1}{|c|}{\textbf{MNIST}}     & 0.0998                                   & 0.0924                             \\ \hline
\multicolumn{1}{|c|}{\textbf{Fashion-MNIST}}     & 0.2890                                  & 0.2835                             \\ \hline
\end{tabular}
\end{table}

\subsubsection{Network Setup}
In this section, we introduce the network structures and settings used to run the experiments on different test cases. In all the test cases, we set epoch = 100 for training autoencoders on original training data. To train the autoencoders on VBD, we set epoch= 10 in all test cases except MNIST datasets. The architecture of trained autoencoders per each dataset are tabulated in Table 1. In all cases, we used \textit{relu} activation function in all layers and \textit{sigmoid} activation function at the last layer. To implement the Variational Autoencoder we used a code which is available in [45] and the repository is located in [46].
\subsubsection{Validation Error}
 We used only 10 epochs to train the autoencoders by VBD on all datasets except MNIST and Fashion-MNIST. In these datasets, we used Algorithm 1 to generate VBD. Our Experiments show that, even after 10 epochs, the validation loss is significantly lower in autoencoders trained by VBD comparing to trained autoencoders by original datasets with 100 epochs. It shows that, VBD can significantly decrease the validation loss even after 10 epochs as shown in Table 2. In case of MNIST dataset, we tested the validation loss per different VBD size as shown in part (c) of Figure 2.  Our observations show that, as the size of VBD increases, the validation loss is decreased. Note that, the MNIST results shown in Table 2 and parts (a),(b) of Figure 2 are obtained by 300000 instances as VBD.

\begin{table}[]
\small
\centering
\caption{Deep anomaly detection results.}
\begin{tabular}{c|c|c|c|}
\cline{2-4}
                                                    & \textbf{Precision} & \textbf{Recall} & \textbf{F1} \\ \hline
\multicolumn{1}{|c|}{\textbf{WBC (VBD)}}            & 0.6123             & 0.7942          & 0.6915      \\ \hline
\multicolumn{1}{|c|}{\textbf{WBC (Original)}}       & 0.5975             & 0.409           & 0.4856      \\ \hline
\multicolumn{1}{|c|}{\textbf{Pima (VBD)}}           & 0.9251             & 0.343           & 0.5005      \\ \hline
\multicolumn{1}{|c|}{\textbf{Pima (Original)}}      & 0.8283             & 0.3552          & 0.4972      \\ \hline
\multicolumn{1}{|c|}{\textbf{Parkinson (VBD)}}      & 0.9166             & 0.2514          & 0.3946      \\ \hline
\multicolumn{1}{|c|}{\textbf{Parkinson (Original)}} & 0.9166             & 0.2404          & 0.3809      \\ \hline
\multicolumn{1}{|c|}{\textbf{Blood (VBD)}}          & 0.9017             & 0.2367          & 0.375       \\ \hline
\multicolumn{1}{|c|}{\textbf{Blood (Original)}}     & 0.882              & 0.2305          & 0.3655      \\ \hline
\multicolumn{1}{|c|}{\textbf{Haberman (VBD)}}       & 0.962              & 0.2704          & 0.4222      \\ \hline
\multicolumn{1}{|c|}{\textbf{Haberman (Original)}}  & 0.9259             & 0.2659          & 0.4132      \\ \hline
\end{tabular}
\end{table}

\begin{table*}[]
\small
\centering
\caption{Required thresholds for deep anomaly detection.}
\begin{tabular}{c|c|c|c|c|c|}
\cline{2-6}
                                                       & \textbf{WBC} & \textbf{Pima} & \textbf{Parkinson} & \textbf{Blood} & \textbf{Haberman} \\ \hline
\multicolumn{1}{|c|}{\textbf{Threshold (VBD)}}         & 12           & 18            & 17                 & 18             & 19                \\ \hline
\multicolumn{1}{|c|}{\textbf{Threshold (Traditional)}} & 3.3          & 3.5           & 6.8                & 2.8            & 2.1               \\ \hline
\end{tabular}
\end{table*}

\subsubsection{Anomaly Detection}
In this section, we evaluate the anomaly detection method proposed in Algorithm 3. To do so, we consider anomaly detection as a solution for one class classification problem. In this case, we trained an autoencoder by minority instances per each tested dataset. The mission of trained autoencoder as anomaly detector is to classify test data into positive and negative instances. To test the anomaly detection methods, we used traditional classification measures including precision, recall and F1 score which are described at the end of this section. In these experiments, precision means the efficiency of anomaly detection method to detect outliers which are the negative instances in the test data. Recall, represents the efficiency of anomaly detection methods to detect the positive instances in the test data. Note that, we used only positive instances (minorities) to train the autoencoders. Therefore, it's so important to know which one of methods performs well in general to detect positive and negative instances. That's why we also used F1 score as a compound measure to evaluate the anomaly detection methods. Our experiments showed that, the proposed anomaly detection method based on VBD can outperform the traditional deep anomaly detection in tested datasets as shown in Table 3 in terms of all tested measures. Furthermore, the universal threshold setting is much easier from practical point of view. We trained the autoencoders by VBD with three epochs and $u=20$ (number of new versions per each test data). Table 4 shows the thresholds used to detect the outliers using the original and VBD. The VBD threshold ($w$) represents the number of times in which a supposed anomalous test data must get higher reconstruction error comparing to randomly selected training pairs. Manipulating the positive integer thresholds is more convenient than traditional thresholds. No matter what's the data type or what's the reconstruction error, the proposed anomaly detection method based on VBD provides us a universal threshold system which is easier than manipulating a float number.

\subsubsection{Performance Measures}
Classifier performance metrics are typically evaluated by a confusion matrix, as shown in following table.
\begin{table}[H]
\centering
\small{}

\begin{tabular}{|l|l|l|}
\hline
                         & \textbf{Detected Positive} & \textbf{Detected Negative} \\ \hline
\textbf{Actual Positive} & TP                         & FN                         \\ \hline
\textbf{Actual Negative} & FP                         & TN                         \\ \hline
\end{tabular}
\end{table}
The rows are actual classes, and the columns are detected classes. TP (True Positive) is the number of correctly classified positive instances. FN (False Negative) is the number of incorrectly classified
positive instances. FP (False Positive) is the number of incorrectly classified negative instances. TN (True Negative) is the number of correctly classified negative instances. The three performance measures are defined by formulae (1)
through (3). \\
\textbf{Recall} = TP/(TP+ FN), \textbf{(1)} \\
\textbf{Precision} = TP/(TP+ FP), \textbf{(2)} \\
\textbf{F1} = (2* Recall * Precision) /( Recall+ Precision) \textbf{(3)} \\

\subsection{Cross-Concatenation Experiments}
In this section, we demonstrate the superiority and advantages of Cross-Concatenation versus SMOTE as the most popular over-sampling method.

\subsubsection{Classification Results}
In our experiments, we applied 10-fold cross-validation to evaluate precision, recall, F1, and AUC score. We employed the Naive Bayes, Logistic Regression and Neural Networks to compare SMOTE, Border Line SMOTE, ADASYN, Random over-sampling and Cross-Concatenation as shown in Figure3. In case of Blood dataset, we used Naive Bayes as base classifier. The experimental results show the superiority of Cross-Concatenation in terms of all metrics except precision. We used Logistic Regression as base classifier in case of Parkinson dataset. The experimental results prove the superiority of Cross-Concatenation in terms of all metrics except precision. In case of Haberman dataset, the experimental results show the superiority of Cross-Concatenation in terms of all metrics except precision using Naive Bayes as base classifier. To test the Pima dataset, we used Naive Bayes as base classifier. The experimental results show the superiority of Cross-Concatenation in terms of all metrics except precision and AUC. In case of Ionosphere dataset, we used Logistic Regression as base classifier. The experimental results show the superiority of Cross-Concatenation in terms of all metrics except precision. We used Neural Network as base classifier to test the WBC dataset. The experimental results show the superiority of Cross-Concatenation in terms of all metrics as shown in Figure 3.
\subsubsection {Separation  }
We proved that, Cross-Concatenation can create larger margins with more separated classes as discussed in section 4. In order to test this theory,  we need to compare the projection quality of Cross-Concatenation. To do so , we test the original data versus its projected version using the linear SVM as base classifier. The experimental results show that, Cross-Concatenation can project the data into a space with larger margins with better class separation. Table 5 shows that, CC-SVM, a linear SVM trained by Cross-Concatenated data reaches better performance results in terms of all metrics.
\subsubsection {Advantages}
In this section, we summarize the main advantages of Cross-Concatenation versus traditional data driven approaches for imbalanced data classification.
\begin{itemize}
    \item \textbf{Stability}: SMOTE is the most popular over-sampling method. However, its random nature makes the synthesized data and even imbalanced classification results unstable. It means that, in case of running SMOTE \textit{n} different times, \textit{n} different synthesized instances are obtained with \textit{n} different classification results. However, there is no any random process in the Cross-Concatenation. That's why in case of running the Cross-Concatenation on fixed training and test data several times, the same efficiency results are obtained.
    
    \item \textbf{Over-fitting}:  Over-fitting is always considered as an imminent consequence of over-sampling techniques. That's why, the synthesized data are considered with skepticism. However, Cross-Concatenation does not create the synthesized data. Instead, it projects the data into a novel space without possibility of creating redundant data which is the main cause of over-fitting.
\end{itemize}

\begin{table*}[]
\caption{SVM versus CC-SVM (Linear SVM on Cross-Concatenated data.}
\begin{tabular}{ccccccccccc}

\cline{2-3} \cline{6-7} \cline{10-11}
\multicolumn{1}{c|}{\textbf{ionosphere}} & \multicolumn{1}{c|}{\textit{\textbf{SVM}}} & \multicolumn{1}{c|}{\textit{\textbf{CC-SVM}}} & \textbf{}             & \multicolumn{1}{c|}{\textbf{Blood}}     & \multicolumn{1}{c|}{\textit{\textbf{SVM}}} & \multicolumn{1}{c|}{\textit{\textbf{CC-SVM}}} & \textbf{}             & \multicolumn{1}{c|}{\textbf{WBC}}       & \multicolumn{1}{c|}{\textit{\textbf{SVM}}} & \multicolumn{1}{c|}{\textit{\textbf{CC-SVM}}} \\ \cline{1-3} \cline{5-7} \cline{9-11} 
\multicolumn{1}{|c|}{\textbf{Recall}}    & \multicolumn{1}{c|}{0.3175}                & \multicolumn{1}{c|}{0.5352}                   & \multicolumn{1}{c|}{} & \multicolumn{1}{c|}{\textbf{Recall}}    & \multicolumn{1}{c|}{0.765}                 & \multicolumn{1}{c|}{0.7912}                   & \multicolumn{1}{c|}{} & \multicolumn{1}{c|}{\textbf{Recall}}    & \multicolumn{1}{c|}{0.9592}                & \multicolumn{1}{c|}{0.9658}                   \\ \cline{1-3} \cline{5-7} \cline{9-11} 
\multicolumn{1}{|c|}{\textbf{Precision}} & \multicolumn{1}{c|}{0.484}                 & \multicolumn{1}{c|}{0.5068}                   & \multicolumn{1}{c|}{} & \multicolumn{1}{c|}{\textbf{Precision}} & \multicolumn{1}{c|}{0.3666}                & \multicolumn{1}{c|}{0.385}                    & \multicolumn{1}{c|}{} & \multicolumn{1}{c|}{\textbf{Precision}} & \multicolumn{1}{c|}{0.9473}                & \multicolumn{1}{c|}{0.9432}                   \\ \cline{1-3} \cline{5-7} \cline{9-11} 
\multicolumn{1}{|c|}{\textbf{F1}}        & \multicolumn{1}{c|}{0.2833}                & \multicolumn{1}{c|}{0.4369}                   & \multicolumn{1}{c|}{} & \multicolumn{1}{c|}{\textbf{F1}}        & \multicolumn{1}{c|}{0.4927}                & \multicolumn{1}{c|}{0.5245}                   & \multicolumn{1}{c|}{} & \multicolumn{1}{c|}{\textbf{F1}}        & \multicolumn{1}{c|}{0.9524}                & \multicolumn{1}{c|}{0.9535}                   \\ \cline{1-3} \cline{5-7} \cline{9-11} 
\multicolumn{1}{|c|}{\textbf{AUC}}       & \multicolumn{1}{c|}{0.5692}                & \multicolumn{1}{c|}{0.6059}                   & \multicolumn{1}{c|}{} & \multicolumn{1}{c|}{\textbf{AUC}}       & \multicolumn{1}{c|}{0.6777}                & \multicolumn{1}{c|}{0.6944}                   & \multicolumn{1}{c|}{} & \multicolumn{1}{c|}{\textbf{AUC}}       & \multicolumn{1}{c|}{0.9648}                & \multicolumn{1}{c|}{0.9665}                   \\ \cline{1-3} \cline{5-7} \cline{9-11} 
\multicolumn{1}{l}{}                     & \multicolumn{1}{l}{}                       & \multicolumn{1}{l}{}                          & \multicolumn{1}{l}{}  & \multicolumn{1}{l}{}                    & \multicolumn{1}{l}{}                       & \multicolumn{1}{l}{}                          & \multicolumn{1}{l}{}  & \multicolumn{1}{l}{}                    & \multicolumn{1}{l}{}                       & \multicolumn{1}{l}{}                          \\ \cline{2-3} \cline{6-7} \cline{10-11} 
\multicolumn{1}{c|}{\textbf{Haberman}}   & \multicolumn{1}{c|}{\textit{\textbf{SVM}}} & \multicolumn{1}{c|}{\textit{\textbf{CC-SVM}}} & \textbf{}             & \multicolumn{1}{c|}{\textbf{Parkinson}} & \multicolumn{1}{c|}{\textit{\textbf{SVM}}} & \multicolumn{1}{c|}{\textit{\textbf{CC-SVM}}} & \textbf{}             & \multicolumn{1}{c|}{\textbf{Pima}}      & \multicolumn{1}{c|}{\textit{\textbf{SVM}}} & \multicolumn{1}{c|}{\textit{\textbf{CC-SVM}}} \\ \cline{1-3} \cline{5-7} \cline{9-11} 
\multicolumn{1}{|c|}{\textbf{Recall}}    & \multicolumn{1}{c|}{0.5736}                & \multicolumn{1}{c|}{0.5925}                   & \multicolumn{1}{c|}{} & \multicolumn{1}{c|}{\textbf{Recall}}    & \multicolumn{1}{c|}{0.6403}                & \multicolumn{1}{c|}{0.8033}                   & \multicolumn{1}{c|}{} & \multicolumn{1}{c|}{\textbf{Recall}}    & \multicolumn{1}{c|}{0.3175}                & \multicolumn{1}{c|}{0.5352}                   \\ \cline{1-3} \cline{5-7} \cline{9-11} 
\multicolumn{1}{|c|}{\textbf{Precision}} & \multicolumn{1}{c|}{0.3709}                & \multicolumn{1}{c|}{0.3818}                   & \multicolumn{1}{c|}{} & \multicolumn{1}{c|}{\textbf{Precision}} & \multicolumn{1}{c|}{0.7314}                & \multicolumn{1}{c|}{0.6177}                   & \multicolumn{1}{c|}{} & \multicolumn{1}{c|}{\textbf{Precision}} & \multicolumn{1}{c|}{0.484}                 & \multicolumn{1}{c|}{0.5068}                   \\ \cline{1-3} \cline{5-7} \cline{9-11} 
\multicolumn{1}{|c|}{\textbf{F1}}        & \multicolumn{1}{c|}{0.4413}                & \multicolumn{1}{c|}{0.4752}                   & \multicolumn{1}{c|}{} & \multicolumn{1}{c|}{\textbf{F1}}        & \multicolumn{1}{c|}{0.6642}                & \multicolumn{1}{c|}{0.6818}                   & \multicolumn{1}{c|}{} & \multicolumn{1}{c|}{\textbf{F1}}        & \multicolumn{1}{c|}{0.2833}                & \multicolumn{1}{c|}{0.4369}                   \\ \cline{1-3} \cline{5-7} \cline{9-11} 
\multicolumn{1}{|c|}{\textbf{AUC}}       & \multicolumn{1}{c|}{0.6096}                & \multicolumn{1}{c|}{0.6211}                   & \multicolumn{1}{c|}{} & \multicolumn{1}{c|}{\textbf{AUC}}       & \multicolumn{1}{c|}{0.7824}                & \multicolumn{1}{c|}{0.8192}                   & \multicolumn{1}{c|}{} & \multicolumn{1}{c|}{\textbf{AUC}}       & \multicolumn{1}{c|}{0.5692}                & \multicolumn{1}{c|}{0.6059}                   \\ \hline
\end{tabular}
\centering

\end{table*}

\begin{figure*}[]
\centering
  \includegraphics[width=120mm]{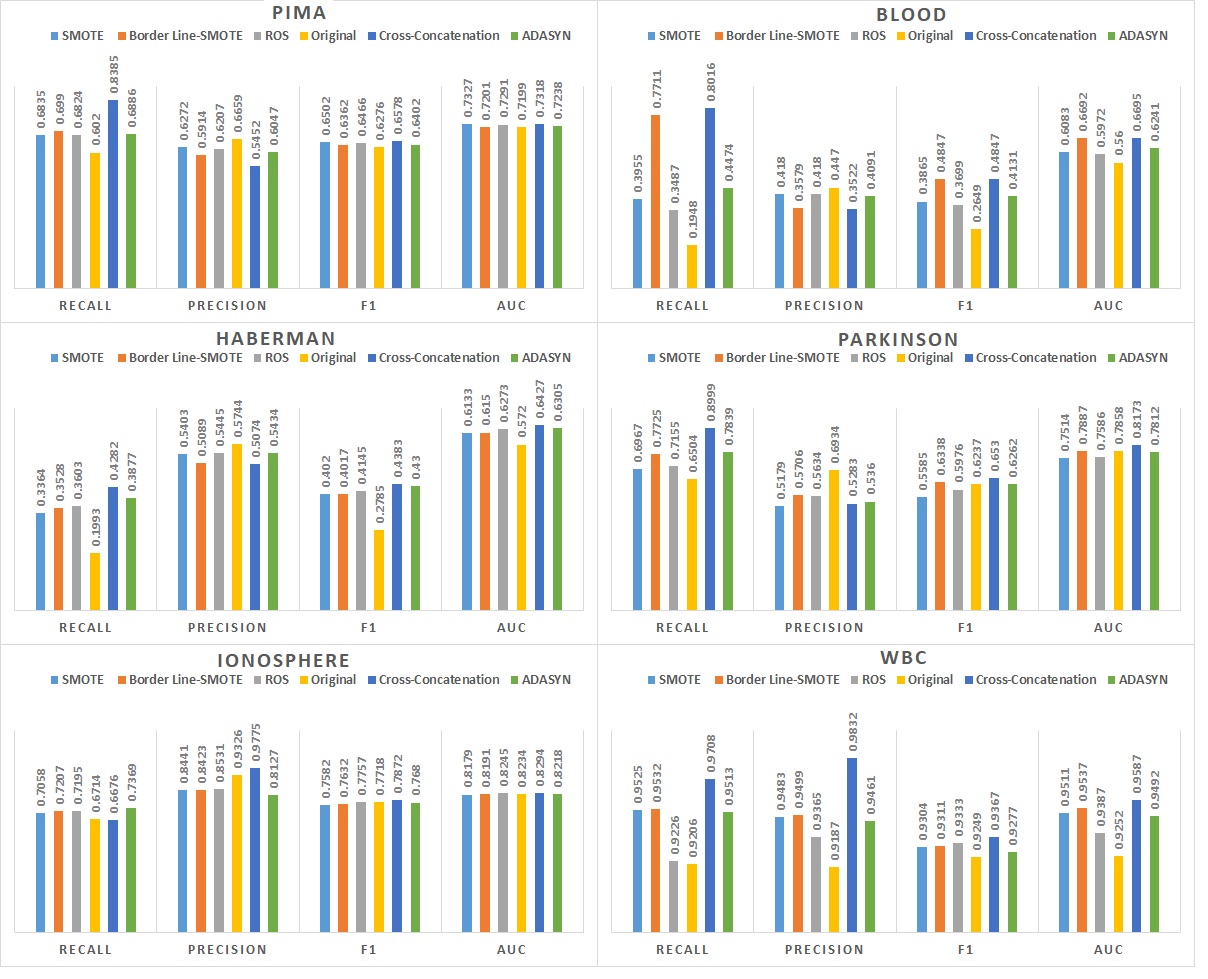}
  \caption{Classification results on six benchmark datasets using Cross-Concatenation.}
  \label{ }
\end{figure*}

\section{Conclusion}
VBD proved its capability to obviate two major problems of the GANs including the mode collapse and diminishing generator gradients very recently. In this paper we showed that, VBD can enhance the efficiency of deep autoencoders. First, we successfully tested VBD to decrease the validation loss of autoencoders on different datasets. Second, we proposed Cross-Concatenation, the first projection-based method to address imbalanced data classification problem using VBD. Cross-Concatenation is the first projection method which can equalize the size of both minority and majority classes. We proved that, Cross-Concatenation can create larger margins with better class separation. Despite SMOTE and its variations, Cross-Concatenation is not based on random procedures. Thus, it can solve instability of over-sampling techniques.

\end{document}